\documentclass[letterpaper,10pt,conference]{ieeeconf}
\IEEEoverridecommandlockouts
\usepackage[T1]{fontenc}
\usepackage{mathptmx}
\usepackage{textcomp}
\usepackage{amsmath,amssymb}
\usepackage{graphicx}
\makeatletter
\def\endfigure{\end@float}
\def\endtable{\end@float}
\makeatother
\usepackage[caption=false,font=footnotesize,labelfont=bf,textfont=rm,farskip=3pt,captionskip=3pt]{subfig}
\usepackage{booktabs,tabularx,makecell,multirow}
\usepackage{cite}
\usepackage{placeins}
\usepackage[hidelinks]{hyperref}
\usepackage{url}

\newcommand{\meansd}[2]{#1\,{\scriptsize\textpm\,#2}}
\newcommand{\bestmeansd}[2]{\textbf{#1}\,{\scriptsize\textpm\,#2}}
\newcommand{\best}[1]{\textbf{#1}}

\newcommand{\tablenote}[1]{\par\smallskip{\footnotesize\normalfont\rmfamily\raggedright\noindent #1\par}}
\newsavebox{\resulttablebox}
\newcommand{\resulttablestyle}{\small\normalfont\rmfamily\renewcommand{\arraystretch}{1.08}}

\newif\ifanonymous
\anonymousfalse
\title{\LARGE\bfseries G$^3$VLA: Geometric Inductive Bias for Vision-Language-Action Models}
\author{Yue Peng$^{1}$, Yongzhe Zhao$^{1}$, Artur Habuda$^{2}$, Khuyen Pham$^{3}$,\\
Yanheng Zhu$^{1}$, Tran Nguyen Le$^{\dagger,2}$, Fares Abu-Dakka$^{\dagger,4}$, and Li Guo$^{\dagger,1}$%
\thanks{\textit{(Yue Peng and Yongzhe Zhao are co-first authors.)}}%
\thanks{$^{\dagger}$Project leads.}%
\thanks{$^{1}$Yue Peng, Yongzhe Zhao, Yanheng Zhu, and Li Guo are with New York University Shanghai, Shanghai, China.}%
\thanks{$^{2}$Artur Habuda and Tran Nguyen Le are with Technical University of Denmark, Denmark.}%
\thanks{$^{3}$Khuyen Pham is with Mohamed bin Zayed University of Artificial Intelligence (MBZUAI), Abu Dhabi, United Arab Emirates.}%
\thanks{$^{4}$Fares Abu-Dakka is with REACH-Lab., Mechanical Engineering Program, Division of Engineering, New York University Abu Dhabi, Abu Dhabi, United Arab Emirates.}%
}
\hypersetup{pdftitle={G3VLA: Geometric Inductive Bias for Vision-Language-Action Models},pdfauthor={Yue Peng, Yongzhe Zhao, Artur Habuda, Khuyen Pham, Yanheng Zhu, Tran Nguyen Le, Fares Abu-Dakka, Li Guo}}

\begin{document}
\raggedbottom
\maketitle
\thispagestyle{empty}
\pagestyle{empty}

\begin{abstract}
Vision-language-action (VLA) models have rapidly advanced
generalist robot manipulation by harnessing the semantic
knowledge of pretrained vision-language backbones.
However, their visual representations are typically grounded
in 2D image coordinates rather than calibrated camera
geometry. This limitation is especially pronounced in
multi-camera systems, where views are coupled by known
intrinsics and extrinsics but processed as independent
images. We propose G$^3$VLA, a camera-aware geometric
module that injects calibrated structure into the
visual-token stream of a pretrained VLA without altering
its action space or original action-learning objective.
G$^3$VLA combines intrinsic-conditioned ray embeddings,
projective positional encoding (PRoPE), and bidirectional
cross-view fusion to establish geometrically consistent
representations across camera views. Geometric supervision
is provided by ground-truth point maps or confidence-gated
$\pi^3$X teacher predictions, enabling training without
depth sensors or manual geometric annotations.
On $\pi_0$, G$^3$VLA consistently improves performance
on LIBERO and RoboCasa24 under both supervision regimes,
while ground-truth supervision improves five-task
RoboTwin2.0 averages with three cameras.
Experiments on camera-shifted LIBERO-Goal and real robots
demonstrate improved robustness to unseen camera viewpoints.
Controlled calibration studies across $\pi_0$ and
$\pi_{0.5}$ show that these gains depend on correct camera
geometry while remaining stable under mild calibration
perturbations. These results demonstrate that grounding
VLA visual tokens in calibrated camera geometry provides
a simple and general mechanism for improving geometric
consistency, multi-view reasoning, and viewpoint robustness
in pretrained robot policies.
\end{abstract}

\begin{figure*}[t]
    \centering
    \includegraphics[width=\textwidth,trim=0 65 0 65,clip]{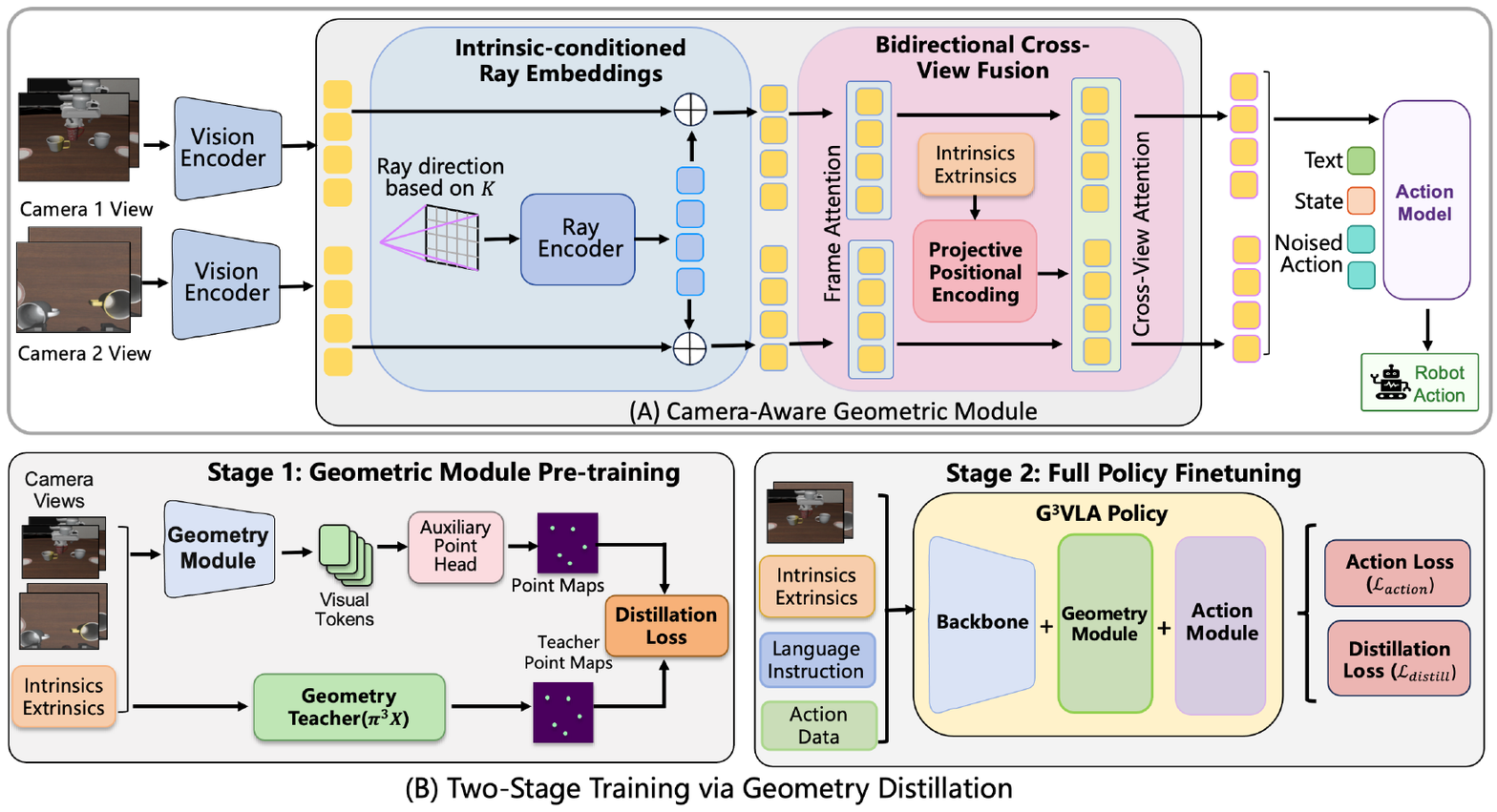}
    \caption{ \textbf{G$^3$VLA overview.} (A) Geometric inductive bias is injected into VLA visual tokens via intrinsic-conditioned ray embeddings ($K^{-1}$) and bidirectional cross-view fusion with PRoPE, preserving the backbone architecture and action objective. (B) Stage 1 uses dense point-map supervision to pretrain the geometry modules; Stage 2 fine-tunes the full policy under action and distillation losses jointly.}
    \label{fig:architecture}
\end{figure*}

\section{Introduction}

Generalist robot manipulation requires two capabilities that have historically
been difficult to combine: semantic understanding of instructions and objects,
and spatial precision in estimating poses, distances, and cross-view geometry.
Vision-Language-Action (VLA) models such as RT-2, OpenVLA, $\pi_0$, and GR00T
leverage pretrained vision-language backbones to support language-conditioned
control across diverse tasks~\cite{zitkovich2023rt2, kim2024openvla,
black2025pi0, bjorck2025gr00t}, but their visual interfaces remain largely based on 2D-image-token, forcing camera geometry to be learned
implicitly from action supervision. This limitation is especially consequential
in calibrated multi-camera systems, where known intrinsics and extrinsics
geometrically couple the views, yet conventional VLAs process them as independent images. 
This also creates a vulnerability to camera viewpoint shift: moving the camera changes the image-space projection of the same physical scene, while conventional VLAs must infer this geometry from demonstrations despite known camera calibration.

A natural way to address this geometric gap is to make the policy explicitly 3D-aware. Existing approaches improve spatial manipulation through dedicated 3D representations, spatial pretraining, or geometry-aware policy interfaces~\cite{shridhar2023peract,goyal2023rvt,
gervet2023act3d,qu2025spatialvla,zhen2024threedvla}.
However, calibrated robotic systems already provide camera intrinsics and extrinsics that directly encode how image observations relate to physical viewing geometry. This motivates a simpler question: can known camera calibration serve as a geometric inductive bias within the visual-token pathway of pretrained VLAs, improving spatial manipulation and generalization to unseen camera viewpoints? 

We introduce \textbf{G$^3$VLA}, which leverages camera intrinsics and extrinsics to make the visual-token pathway explicitly geometry-aware. G$^3$VLA augments visual tokens with \textit{intrinsic-conditioned ray embeddings}~\cite{zhang2024cameras}, which encode per-patch viewing directions from $K^{-1}$, and models calibrated cross-view relationships through \textit{projective positional encoding} (PRoPE)~\cite{li2025cameras} and \textit{bidirectional cross-view fusion}. Together, these modules expose both per-view and cross-view geometry to the action model while preserving its original action representation. Geometric supervision is provided either by ground-truth point maps when depth is available or by confidence-gated $\pi^3$X~\cite{wang2025pi3} distillation otherwise, requiring no depth sensors or manual annotations.

We evaluate G$^3$VLA across simulation and real-robot manipulation, with particular emphasis on generalization to unseen camera viewpoints. On standard benchmarks, G$^3$VLA consistently improves $\pi_0$ across LIBERO, RoboCasa24, and RoboTwin2.0. On camera-shifted LIBERO-Goal, G$^3$VLA further improves performance under held-out camera viewpoints, with larger gains under more challenging viewpoint changes. Real-robot experiments further evaluate generalization to held-out camera viewpoints. Controlled calibration experiments show that these gains depend on correct camera geometry rather than additional model capacity alone. Finally, experiments on $\pi_{0.5}$ and GR00T~1.5 examine whether the same geometric interface transfers across VLA backbones, showing gains on the near-saturated $\pi_{0.5}$ LIBERO benchmark and more mixed effects on GR00T~1.5. 

Our contributions are threefold: \textbf{(1) Camera geometry as an inductive bias.} We identify known camera calibration as an underexplored geometric prior for pretrained VLAs and study its role in spatial manipulation and generalization to unseen camera viewpoints. \textbf{(2) G$^3$VLA: a geometry-aware visual-token interface.} We introduce a lightweight interface that injects camera intrinsics and extrinsics directly into the visual-token stream through intrinsic-conditioned ray embeddings, projective positional encoding, and bidirectional cross-view fusion, while preserving the pretrained VLA's original action space and learning objective. \textbf{(3) Viewpoint generalization and geometric attribution.} We comprehensively evaluate G$^3$VLA across simulation and real-robot experiments, showing that explicitly grounding visual tokens in calibrated camera geometry improves manipulation performance and robustness to unseen viewpoints. Controlled calibration studies across multiple VLA backbones, multi-camera configurations further isolate correct camera geometry as the source of these gains.

\section{Related Work}

\textbf{Vision-language-action models.}
Pretrained VLA models such as OpenVLA, $\pi_0$, and $\pi_{0.5}$ transfer vision-language knowledge to robot manipulation~\cite{kim2024openvla,black2025pi0,black2025pi05}. Their visual-token interfaces, however, do not explicitly incorporate camera intrinsics and extrinsics.

\textbf{Structured 3D representations for manipulation.}
Many policies improve spatial precision through explicit 3D scene representations: PerAct uses voxelized RGB-D observations~\cite{shridhar2023peract}, RVT aggregates rendered multi-view features~\cite{goyal2023rvt}, Act3D lifts 2D features into a 3D feature field~\cite{gervet2023act3d}, PolarNet operates on point clouds~\cite{chen2023polarnet}, and 3D Diffuser Actor pairs 3D scene representations with action diffusion for stronger cross-viewpoint generalization~\cite{ke2024diffuseractor}. These approaches use explicit 3D scene representations, whereas G$^3$VLA conditions RGB visual tokens on camera calibration.

\textbf{Spatially aware VLAs.}
Other work connects VLA policies more directly to spatial representations: 3D-VLA couples action prediction with 3D world modeling~\cite{zhen2024threedvla}, SpatialVLM studies metric spatial reasoning in Vision-Language Models (VLMs)~\cite{chen2024spatialvlm}, and 3DS-VLA builds a spatially aware VLA around 3D representations~\cite{li2025threedsvla}. Whereas SpatialVLA~\cite{qu2025spatialvla} couples egocentric 3D encodings with adaptive action grids, G$^3$VLA introduces known camera intrinsics and extrinsics into cross-view visual-token fusion while retaining the pretrained policy's action representation.

\textbf{Camera-aware representations.}
PRoPE encodes calibrated cross-view relationships within attention~\cite{li2025cameras}. Visual-geometry models such as DUSt3R, VGGT, and $\pi^3$ predict dense point map from RGB, which can serve as auxiliary geometric supervision~\cite{wang2024dust3r,wang2025vggt,wang2025pi3}. While Spatial Forcing~\cite{spatialforcing} transfers spatial knowledge
through training-time feature alignment, G$^3$VLA complements geometric
supervision with explicit camera calibration conditioning that remains
active during inference.

\section{Method}
\label{sec:method}

We frame the problem as language-conditioned manipulation from calibrated multi-camera observations.  At each control step the policy receives a language
instruction~$l$, proprioceptive state~$s_t$, and RGB images
$\{I_t^v\}_{v=1}^{V}$ from $V$ views, each with intrinsic matrix~$K^v$ and
camera-to-world pose~$T^v$, and predicts an action chunk:
\begin{equation}
    \pi_\theta(a_{t:t+H-1}\mid l,\, s_t,\, \{I_t^v, K^v, T^v\}_{v=1}^{V}).
\end{equation}

\textbf{G$^3$VLA} uses this calibration by inserting a \emph{Camera-Aware
Geometric Module} into the visual-token stream before action prediction
(Fig.~\ref{fig:architecture}), leaving the backbone architecture, action space,
and original action loss unchanged.

\subsection{Camera-Aware Geometric Module}
\label{sub:geometric_module}

The patch-embedding layer of a pretrained vision
encoder~\cite{dosovitskiy2020image, zhai2023sigmoid} produces $P$ tokens
per view, $e_p^v \in \mathbb{R}^d$. Our geometry-aware visual pathway
conditions their encoding and cross-view fusion on camera calibration:
\begin{equation}
    h_{1:P}^{1:V}
    = F_\psi\!\left(e_{1:P}^{1:V},\;\{K^v,T^v\}_{v=1}^{V}\right).
\end{equation}

As shown in Fig.~\ref{fig:architecture}(A), $F_\psi$ adds
intrinsic-conditioned ray embeddings before the vision Transformer,
followed by PRoPE-guided bidirectional cross-view fusion.

\textbf{Intrinsic-conditioned Ray Embeddings.}
Under different intrinsics the same pixel $(x,y)$ maps to a different physical
viewing direction---an ambiguity that 2D positional embeddings cannot resolve.
For a homogeneous pixel coordinate $u=(x,y,1)^\top$, the normalized projection
ray is $\tilde{r}^{\,v}(u) = (K^v)^{-1}u$.  Since the pinhole model fixes the
third coordinate, the first two components define an image-plane ray coordinate
without assuming metric depth:
\begin{equation} \label{eq:ray}
    R^v(x,y)
    = \bigl[\tilde{r}^{\,v}(x,y,1)\bigr]_{1:2}
    \in \mathbb{R}^{2}.
\end{equation}
A learnable embedding $G_\phi$ projects this ray map to the patch grid and adds
it to the patch embeddings before the SigLIP Transformer encoder:
\begin{equation} \label{eq:rayadd}
    z_{0,p}^v = e_p^v + G_\phi(R^v)_p + b_p,
\end{equation}
where $b_p$ is the pretrained positional embedding. The encoder processes
$z_{0,1:P}^v$ into visual tokens for cross-view fusion, so camera-local
ray information is available from the first encoder attention layer.

\textbf{Projective Positional Encoding (PRoPE).}
Ray embeddings encode the viewing direction local to each camera but do not
capture how tokens from \emph{different} views relate geometrically.
PRoPE~\cite{li2025cameras} fills this gap by deriving fixed projective
transforms for the query, key, and value representations from per-view
intrinsics~$K^v$, world-to-camera matrices~$(T^v)^{-1}$, and patch
locations---giving cross-view attention access to camera-model-based projective
relations rather than relying on appearance similarity alone.

\textbf{Bidirectional Cross-View Fusion.}
The fusion module proceeds in two steps.  \emph{Frame Attention} processes
tokens within each camera stream independently, preserving view-local structure.
\emph{Cross-View Attention} then flattens view and patch dimensions so that all
valid tokens attend bidirectionally across views. This stage comprises a
standard global attention block followed by a separate PRoPE block whose
output is added residually to the tokens. The fused tokens retain the
pretrained VLA's visual-token interface and are passed to the action model.

Using the PRoPE-transformed queries $\widetilde Q$, keys $\widetilde K$,
and values $\widetilde V$, the PRoPE block computes the intermediate aggregate
\begin{equation}
    A = \operatorname{softmax}\!\left(
        \widetilde Q\widetilde K^\top/\sqrt{d_h} + M
    \right)\widetilde V,
    \label{eq:prope_attention}
\end{equation}
where $d_h$ is the attention-head dimension and $M$ masks invalid or padded
views. A corresponding per-token PRoPE output transform is applied to $A$
before the output projection, conditioning both attention weights and value
aggregation on relative camera geometry. The same operation accommodates all valid camera streams without
requiring explicit correspondences. For moving cameras, the extrinsics are
updated at each observation.

\textbf{Implementation.}
We use $224\times224$ inputs with $16\times16$ SigLIP patch grids.
The ray projection $G_\phi$ is zero-initialized, and frame attention
precedes one global cross-view block and one PRoPE block. Camera-to-world poses are
inverted to world-to-camera matrices for PRoPE. We use neither
camera-ID embeddings nor calibration-inconsistent augmentations. \looseness=-1
\subsection{Geometric Supervision and Two-Stage Training}
\label{sub:training}

Action supervision alone does not explicitly constrain the geometry encoded
by the new visual-token module. We therefore introduce a dense
auxiliary \emph{geometry distillation} objective and a \emph{two-stage training
curriculum} to bootstrap the geometric pathway before full policy fine-tuning.

\textbf{Geometric Supervision Objective.}
An auxiliary point head, attached to the fused visual tokens before VLA
projection, produces per-pixel predictions
$\hat{q}_u^v\!\in\!\mathbb{R}^2$ (ray coordinate) and
$\hat{d}_u^v\!\in\!\mathbb{R}$ (log-$z$ depth).
Together, the ray coordinate and log-$z$ depth parameterize a 3D point
in the source camera frame.
Supervision targets $q_u^v$ and $d_u^v$ come from one of two sources.  When
ground-truth depth is available---e.g.\ in simulation---we set the validity
mask $m_u^v=1$ at valid depth pixels. Otherwise, we use
$\pi^3$X~\cite{wang2025pi3} as an offline geometry teacher, obtaining
per-pixel targets and confidence score~$c_u^v$ converted to a hard gate:
\begin{equation} \label{eq:gate}
    m_u^v = \mathbf{1}\!\left[c_u^v > \tau\right],
\end{equation}
with $\tau=0.1$.  The distillation loss is unified across both modes:
\begin{equation} \label{eq:distill}
    \mathcal{L}_{\mathrm{distill}}
    = \frac{
        \displaystyle\sum_{v,u} m_u^v
        \!\left(
            \tfrac{1}{2}\lVert \hat{q}_u^v - q_u^v \rVert_2^2
            + (\hat{d}_u^v - d_u^v)^2
        \right)
    }{
        \displaystyle\sum_{v,u} m_u^v + \epsilon
    }.
\end{equation}

For $\pi^3$X supervision, converted LeRobot frames are processed
independently at $224\times224$, with intrinsics scaled accordingly.
We cache the two ray-coordinate channels, log-$z$ and confidence logits, where $z$ is the depth along the source camera's optical axis, kept at the teacher's scale.

For simulator supervision, rendered depth is transformed using the
same image rotation and resolution, with
$q=((x-c_x)/f_x,(y-c_y)/f_y)$ and $d=\log z$. Invalid or missing
depth is gated out. Both target sources use the same auxiliary head,
full-resolution loss in Eq.~\eqref{eq:distill}, and confidence gate.

\textbf{Two-Stage Training Curriculum.}
We combine the action and distillation losses as
\[
    \mathcal{L}
    = \lambda_{\mathrm{act}}\mathcal{L}_{\mathrm{act}}
    + \lambda_{\mathrm{distill}}\mathcal{L}_{\mathrm{distill}},
\]
where $\mathcal{L}_{\mathrm{act}}$ is the base VLA's action
objective---the flow-matching loss for $\pi_0$. We train Stages 1
and 2 for 5k and 30k steps with 500 and 1k warmup steps and
$(\lambda_{\mathrm{act}},\lambda_{\mathrm{distill}})
=(0.1,1.0)$ and $(1.0,0.05)$, respectively. Both stages use AdamW
($\beta_1=0.9,\beta_2=0.95,\epsilon=10^{-8}$), negligible weight
decay, global gradient clipping at 1.0, global batch size 32,
bfloat16 precision, and cosine learning-rate decay from
$2.5\times10^{-5}$ to $2.5\times10^{-6}$.
\textbf{Stage~1: Geometric Module Pretraining.}
We update only the ray embeddings, cross-view fusion layers, and auxiliary point
head, keeping the pretrained backbone frozen and the distillation loss dominant.
\textbf{Stage~2: Full Policy Fine-tuning.}
Starting from the Stage~1 checkpoint, we unfreeze all parameters and let the
action loss dominate, retaining distillation as a lightweight geometric
regularizer. At inference, the auxiliary head is discarded and the teacher
is not queried.
The auxiliary head reshapes each view's 256 fused tokens to a
$16\times16$ grid, processes them with two 8-head Transformer
blocks, and upsamples to $224\times224$, where two zero-initialized
heads predict ray coordinates and log-$z$.

\section{Experiments}
\label{sec:results}

Our experiments address three questions:
(1) Does G$^3$VLA improve manipulation performance across benchmarks and VLA backbones?
(2) Does explicit camera geometry improve generalization to unseen viewpoints?
(3) Are the gains attributable to meaningful geometric conditioning and the proposed design choices?
We answer these questions through standard benchmark evaluation (Sec.~\ref{sec:main_sim_results}),
viewpoint extrapolation (Sec.~\ref{sec:generalization}), and calibration controls and ablations (Secs.~\ref{sec:calibration_controls} and~\ref{sec:ablation_studies}).

\subsection{Experimental Setup}
\label{sec:exp_setup}

\textbf{Simulation benchmarks.}
We evaluate on the four LIBERO suites (Goal, Spatial, Object, and
LIBERO-10)~\cite{liu2023libero}, RoboCasa24 household
tasks~\cite{nasiriany2024robocasa}, and five RoboTwin2.0 bimanual tasks
under Easy and Hard conditions with three camera streams~\cite{chen2025robotwin}.

\textbf{Base policies and variants.}
Our main experiments use $\pi_0$ across LIBERO, RoboCasa24, and
RoboTwin2.0. We additionally evaluate $\pi_{0.5}$ and GR00T~1.5 on
LIBERO. Geometry supervision is provided either by ground truth (GT) or $\pi^3$X,
as defined in Sec.~\ref{sub:training}. \looseness=-1

\textbf{Simulation evaluation protocol.}
Simulation results are evaluated over three independent runs where applicable. 
For LIBERO, each task is evaluated on 50 rollouts from the official
initial-state distribution. We use suite-specific episode limits
(220/280/300/520 steps for Spatial/Object/Goal/LIBERO-10) to account
for their different task horizons. At each policy query, we execute the first five predicted actions before querying again and use the benchmark-defined success condition. 
For RoboCasa24, policies execute 50-action chunks with a 500-step episode
limit. For RoboTwin2.0, we evaluate only seeds validated as feasible by the benchmark expert policy and use the benchmark-defined success condition.

\textbf{Camera calibration and preprocessing.} 
Across simulation benchmarks, fixed external cameras use constant
extrinsics, while end-effector-mounted cameras recompute their pose
from the live simulator state at every observation. Camera intrinsics
$K$ are fixed.  We standardize all camera coordinates to the OpenCV convention
($x$-right, $y$-down, $z$-forward) and keep image transformations
consistent with the corresponding camera intrinsics throughout data
storage, training, evaluation, and geometry supervision. 

\textbf{Real-world benchmarks.}
For real-world experiments, the bimanual UR5 workbench uses one arm for manipulation and the other to hold the context camera. \textit{Pick-and-Place Test Tube} transfers a tube between holders, while \textit{ Pouring Nut} combines container stacking and pouring. Each task uses 120 training episodes from 10 viewpoints (1--10), with a $10{\times}10$\,cm grid to reproduce object positions under camera shifts. Evaluation uses views 1 and 3 (ID) and held-out views 11--13 (OOD). Intrinsics are captured at initialization, and extrinsics are computed per frame from forward kinematics.

For \textit{Pick-and-Place Test Tube}, holder pairs A1--B1 and A2--B3, with the tube placed in source slot 1 or 3, yield four trials per viewpoint. Trials are scored as grasp failure (0), grasp success with placement failure ($1^*$), or full success (1); both $1^*$ and 1 count as success.

For \textit{Pouring Nut}, nut-container locations A1/A2 and wheel-container locations B2/B6 yield four trials per viewpoint. Trials receive 0 for failure, 0.5 for completed stacking but incomplete pouring, $1^*$ when both stages succeed with some nuts dropped outside, and 1 for full completion; both $1^*$ and 1 count as success.

\subsection{Performance on Standard Manipulation Benchmarks}
\label{sec:main_sim_results}

We first instantiate G$^3$VLA on $\pi_0$ and evaluate whether explicit camera geometry improves manipulation performance under standard benchmark settings. Across LIBERO, RoboCasa24, and RoboTwin2.0, G$^3$VLA consistently improves the $\pi_0$ baseline, spanning tabletop, household, and three-camera bimanual manipulation. 

\textbf{LIBERO.}
G$^3$VLA improves the $\pi_0$ macro-average success rate from 84.6\% to 88.1\% with GT geometric supervision and to 87.0\% with $\pi^3$X supervision (Table~\ref{tab:libero}). The $\pi^3$X-supervised variant improves over the baseline across all four suites, showing that G$^3$VLA remains effective without ground-truth geometry. \looseness=-1

\textbf{RoboCasa24.}
G$^3$VLA also improves performance on the more diverse RoboCasa24 household tasks (Table~\ref{tab:broader}). The overall success rate increases from 34.2\% to 37.1\% with GT supervision and to 36.5\% with $\pi^3$X supervision.  

\begin{table}[tbp]
\centering
\resulttablestyle
\setlength{\tabcolsep}{7pt}
\begin{lrbox}{\resulttablebox}
\begin{tabular}{@{}lccc@{}}
\toprule
\multirow{2}{*}{\textbf{Suite}} & \multirow{2}{*}{$\pi_0$} & \multicolumn{2}{c}{\textbf{G$^3$VLA}} \\
\cmidrule(l){3-4}
& & $\pi^3$X & GT \\
\midrule
Goal & \meansd{87.4}{0.5} & \bestmeansd{88.4}{0.6} & \bestmeansd{88.4}{0.4} \\
Spatial & \meansd{85.2}{0.7} & \meansd{88.6}{0.6} & \bestmeansd{89.2}{0.5} \\
Object & \meansd{89.4}{0.5} & \meansd{93.4}{0.5} & \bestmeansd{94.4}{0.4} \\
L-10 & \meansd{76.5}{0.9} & \meansd{77.6}{0.8} & \bestmeansd{80.4}{0.7} \\
\midrule
\textbf{Average} & \meansd{84.6}{0.4} & \meansd{87.0}{0.4} & \bestmeansd{88.1}{0.3} \\
Gain (pp) & --- & +2.4 & \textbf{+3.5} \\
\bottomrule
\end{tabular}
\end{lrbox}
\begin{minipage}{\wd\resulttablebox}
\caption{LIBERO results on $\pi_0$}
\label{tab:libero}
\usebox{\resulttablebox}
\tablenote{Success (\%): mean $\pm$ standard deviation over three evaluation runs. Gain is relative to $\pi_0$ in percentage points (pp); bold marks the highest mean in each row.\looseness=-1}
\end{minipage}
\end{table}

\begin{table}[tbp]
\centering
\resulttablestyle
\setlength{\tabcolsep}{4pt}
\begin{lrbox}{\resulttablebox}
\begin{tabular}{@{}lccc@{}}
\toprule
\multirow{2}{*}{\textbf{Task family}} & \multirow{2}{*}{$\pi_0$} & \multicolumn{2}{c}{\textbf{G$^3$VLA}} \\
\cmidrule(l){3-4}
& & $\pi^3$X & GT \\
\midrule
Pick \& Place & 13.1 & 13.1 & \best{18.1} \\
Doors / Drawers & 53.3 & 52.5 & \best{54.2} \\
Others & 39.5 & \best{45.5} & 42.0 \\
\midrule
\textbf{Overall} & \meansd{34.2}{0.9} & \meansd{36.5}{1.1} & \bestmeansd{37.1}{1.0} \\
Gain (pp) & --- & +2.3 & \textbf{+2.9} \\
\bottomrule
\end{tabular}
\end{lrbox}
\begin{minipage}{\wd\resulttablebox}
\caption{RoboCasa24 results by task family}
\label{tab:broader}
\label{tab:appendix_robocasa_breakdown}
\usebox{\resulttablebox}
\tablenote{Family entries are success rates (\%). Overall is mean $\pm$ standard deviation over three evaluation runs, with 480 trials per run. Gain is relative to $\pi_0$ in percentage points (pp). Bold marks the highest rate in each row.}
\end{minipage}
\end{table}

\textbf{RoboTwin2.0.}
We further evaluate G$^3$VLA on five bimanual tasks with three camera streams under both Easy and Hard conditions (Table~\ref{tab:robotwin_multitask}). With GT supervision, G$^3$VLA improves performance in nine of the ten task--difficulty combinations and matches the baseline in the remaining one, increasing the average success rate from 63.8\% to 69.2\% on Easy and from 11.6\% to 15.8\% on Hard.

\begin{table}[tbp]
\centering
\resulttablestyle
\setlength{\tabcolsep}{3.5pt}
\begin{lrbox}{\resulttablebox}
\begin{tabular}{@{}lcccc@{}}
\toprule
\multirow{2}{*}{\textbf{Task}} & \multicolumn{2}{c}{\textbf{Easy}} & \multicolumn{2}{c}{\textbf{Hard}} \\
\cmidrule(lr){2-3}\cmidrule(l){4-5}
& $\pi_0$ & G$^3$VLA & $\pi_0$ & G$^3$VLA \\
\midrule
Beat Block Hammer & 54.0 & \best{58.0} & 3.0 & \best{9.0} \\
Handover Block & 39.0 & \best{55.0} & 4.0 & \best{8.0} \\
Lift Pot & 52.0 & \best{58.0} & 9.0 & \best{14.0} \\
Place Container Plate & 94.0 & \best{95.0} & 21.0 & \best{23.0} \\
Stack Bowls Two & \best{80.0} & \best{80.0} & 21.0 & \best{25.0} \\
\midrule
\textbf{Average} & 63.8 & \textbf{69.2} & 11.6 & \textbf{15.8} \\
Gain (pp) & --- & \textbf{+5.4} & --- & \textbf{+4.2} \\
\bottomrule
\end{tabular}
\end{lrbox}
\begin{minipage}{\wd\resulttablebox}
\caption{Three-camera RoboTwin2.0 results}
\label{tab:robotwin_multitask}
\usebox{\resulttablebox}
\tablenote{G$^3$VLA uses GT supervision. Success rates (\%) are averaged over three evaluation runs; Average is the five-task macro-average. Bold marks the best scores within each difficulty.}
\end{minipage}
\end{table}

\subsection{Generalization to Unseen Camera Viewpoints}
\label{sec:generalization}

\begin{figure}[t]
    \centering
    \includegraphics[width=\columnwidth,trim=0bp 225bp 0bp 225bp,clip]{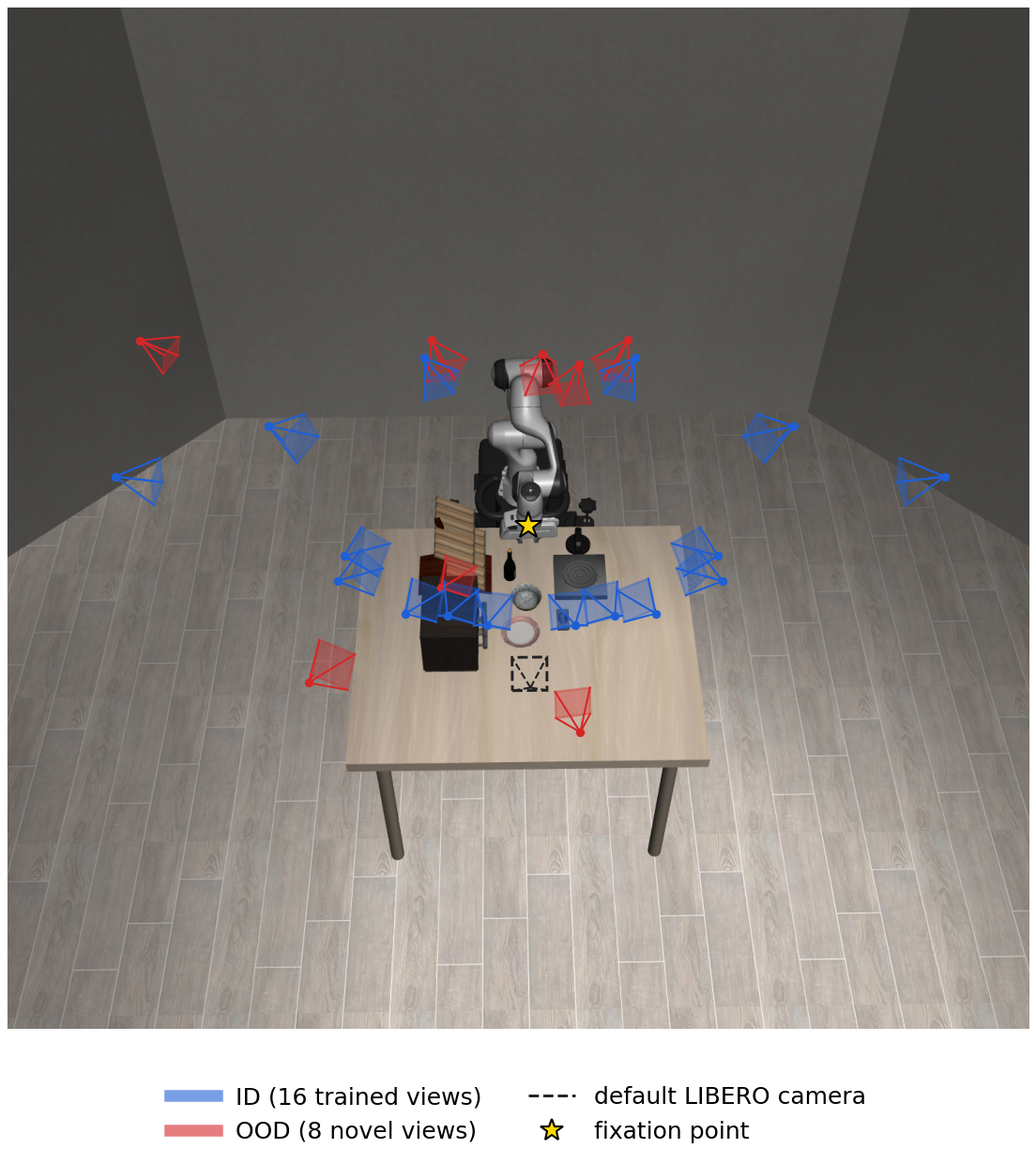}\par
    \vspace{3pt}
    \includegraphics[width=\columnwidth,trim=0bp 7.5bp 0bp 847.5bp,clip]{imgs/cameras.png}
    \caption{\textbf{Camera viewpoints for the camera-shifted LIBERO-Goal dataset.}
        Blue frustums denote the 16 in-distribution (ID) training viewpoints, while red
        frustums denote the 8 held-out out-of-distribution (OOD) viewpoints used to evaluate
        viewpoint extrapolation.}
    \label{fig:camera_viewpoints}
\end{figure}

\textbf{Viewpoint generalization in simulation.}
We construct a camera-shifted LIBERO-Goal benchmark to evaluate
generalization beyond the camera configurations observed during training.
Each of the 500 demonstrations is re-rendered from one of 16 third-person
camera viewpoints spanning variations in yaw, pitch, and radius, while
preserving the original dataset size (Fig.~\ref{fig:camera_viewpoints}).
We train both the baseline $\pi_0$ and $\pi^3$X-supervised G$^3$VLA on
these 16 viewpoints and evaluate them on the training-view distribution
(ID) and 8 held-out viewpoints (OOD).

We evaluate the 20K, 25K, and 30K checkpoints and report both
checkpoint-level and average performance in Table~\ref{tab:camera_shift}.
G$^3$VLA consistently outperforms the baseline at all three checkpoints,
with average success increasing from 86.1\% to 89.1\% on ID viewpoints
and from 80.6\% to 87.0\% on held-out OOD viewpoints. The larger OOD
gain (+6.4 points vs.\ +3.0 on ID) indicates that geometry conditioning
is particularly beneficial under viewpoint extrapolation. Moreover,
G$^3$VLA outperforms the baseline on all eight held-out viewpoints
(Table~\ref{tab:camera_shift_ood_types}), showing that the improvement
extends across different types of camera shift rather than being driven
by a subset of viewpoints. We did not check target visibility frame by frame. In some low or close viewpoints the manipulated object or goal can be temporarily partially outside the third-person frame, in which case the policy must rely on the wrist camera. Excluding such (view, task) pairs does not change the conclusion (ID +2.4, OOD +7.3).
\looseness=-1

\begin{figure*}[t]
    \centering
    \includegraphics[width=0.9\textwidth]{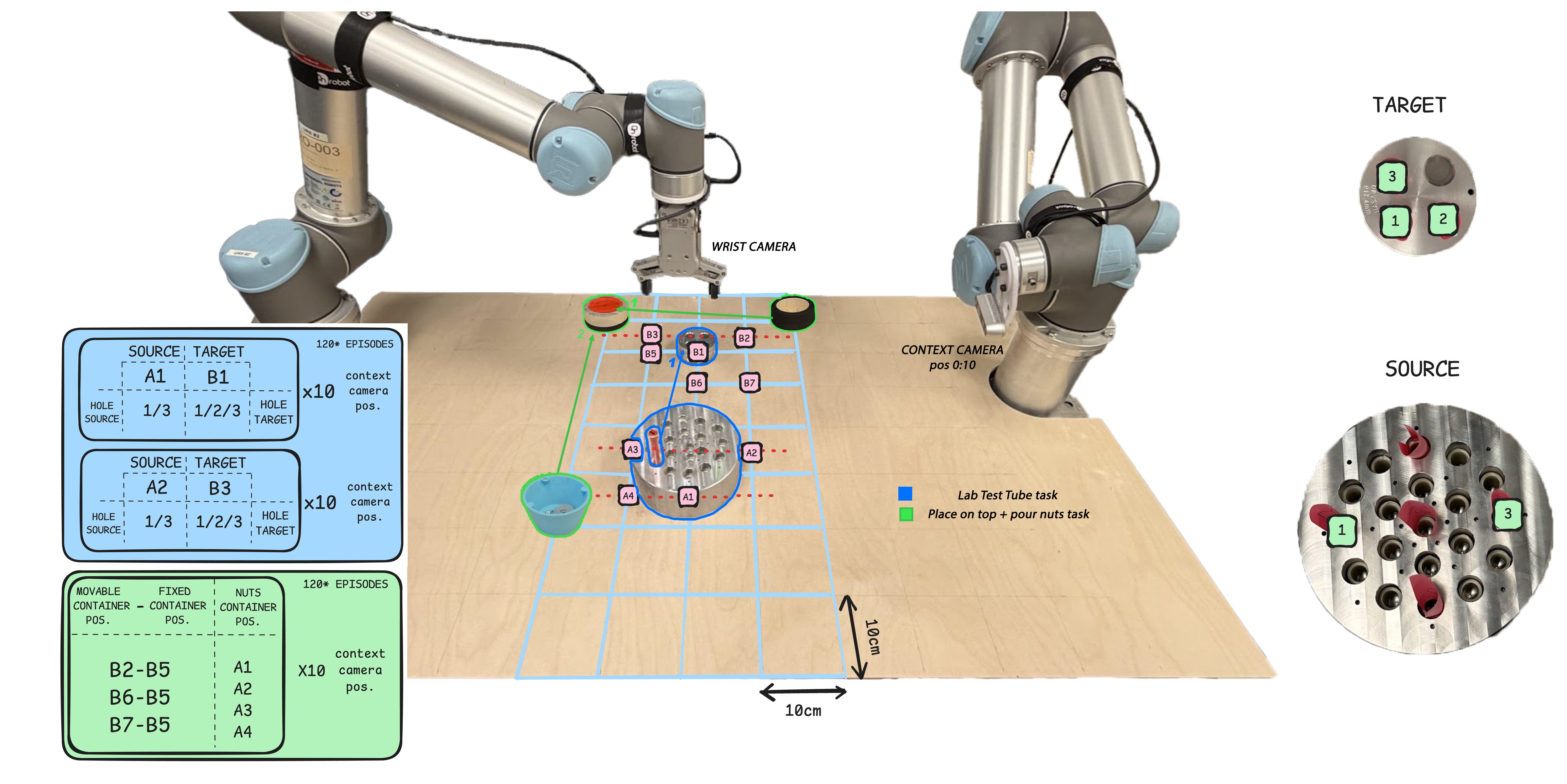}
    \caption{Real-World experimental setup on bimanual UR5 workbench. Two tasks are used for evaluation: \textbf{Pick-and-Place Test Tube} (in blue), and \textbf{Pouring Nut} (in green). Dataset distribution is presented in the table defined by the relevant object positions for a given task. Each combination of positions is repeated 10 times with a different context camera position.}
    \label{fig:real-robo}
\end{figure*}

\begin{table}[tbp]
\centering
\caption{Camera-shifted LIBERO-Goal results}
\label{tab:camera_shift}
\setlength{\tabcolsep}{4.5pt}
\renewcommand{\arraystretch}{1.0}
\begin{tabular}{lcc@{\hspace{8pt}}cc}
\toprule
& \multicolumn{2}{c}{\textbf{ID}} 
& \multicolumn{2}{c}{\textbf{OOD}} \\
\cmidrule(lr){2-3}\cmidrule(lr){4-5}
\textbf{Checkpoint}
& Baseline & G$^3$VLA
& Baseline & G$^3$VLA \\
\midrule
20K
& 85.9 & \textbf{89.0}
& 81.0 & \textbf{87.0} \\

25K
& 86.6 & \textbf{89.2}
& 80.5 & \textbf{87.5} \\

30K
& 85.8 & \textbf{89.2}
& 80.5 & \textbf{86.7} \\

\midrule
\textbf{Average}
& 86.1 & \textbf{89.1}
& 80.6 & \textbf{87.0} \\
Gain (pp) & --- & \textbf{+3.0} & --- & \textbf{+6.4} \\
\bottomrule
\end{tabular}

\vspace{2pt}
\parbox{\columnwidth}{\footnotesize
Success rates (\%) on 16 trained (ID) and 8 held-out (OOD)
viewpoints. Gain is relative to $\pi_0$ in percentage points (pp).
}
\end{table}

\begin{table}[tbp]
\centering
\resulttablestyle
\setlength{\tabcolsep}{4pt}
\begin{lrbox}{\resulttablebox}
\begin{tabular}{@{}lcc@{}}
\toprule
\textbf{OOD viewpoint} & \textbf{Baseline} & \textbf{G$^3$VLA} \\
\midrule
Yaw left$(-110^\circ,\,45^\circ,\,r=0.80)$
    & 78.4 & \textbf{89.4} \\
Yaw right$(+110^\circ,\,45^\circ,\,r=0.80)$
    & 88.5 & \textbf{89.2} \\
Pitch high $(8^\circ,\,55^\circ,\,r=1.02)$
    & 88.5 & \textbf{93.7} \\
Pitch low $(8^\circ,\,-6^\circ,\,r=1.02)$
    & 75.2 & \textbf{81.9} \\
Radius close $(-25^\circ,\,10^\circ,\,r=0.70)$
    & 85.7 & \textbf{86.8} \\
Radius far $(-25^\circ,\,10^\circ,\,r=1.50)$
    & 80.3 & \textbf{91.4} \\
Joint extrapolation $(+110^\circ,\,55^\circ,\,r=0.70)$
    & 80.1 & \textbf{85.6} \\
Joint extrapolation $(-110^\circ,\,-6^\circ,\,r=1.50)$
    & 68.3 & \textbf{78.3} \\
\bottomrule
\end{tabular}
\end{lrbox}
\begin{minipage}{\wd\resulttablebox}
\caption{Success by held-out camera extrapolation viewpoint}
\label{tab:camera_shift_ood_types}
\usebox{\resulttablebox}
\tablenote{Success rates (\%) on individual held-out viewpoints, averaged across checkpoints 20K, 25K, 30K.}
\end{minipage}
\end{table}

\begin{table*}[t]
\centering
\begin{minipage}{0.94\textwidth}
\centering
\resulttablestyle
\setlength{\tabcolsep}{3.8pt}
\caption{Real-world task scores (\%)}
\label{tab:real_robot_res}
\begin{tabular*}{\linewidth}{@{\extracolsep{\fill}}lcccccccccccc@{}}
\toprule
& \multicolumn{3}{c}{\textbf{$\pi_0$}}
& \multicolumn{3}{c}{\textbf{G$^3$VLA ($\pi_0$)}}
& \multicolumn{3}{c}{\textbf{$\pi_{0.5}$}}
& \multicolumn{3}{c}{\textbf{G$^3$VLA ($\pi_{0.5}$)}} \\
\cmidrule(lr){2-4}
\cmidrule(lr){5-7}
\cmidrule(lr){8-10}
\cmidrule(lr){11-13}
\textbf{Chkpt.}
& ID & OOD & Overall
& ID & OOD & Overall
& ID & OOD & Overall
& ID & OOD & Overall \\
\midrule
\multicolumn{13}{l}{\textbf{P\&P Test Tube} \quad G$^3$VLA supervision: $\pi^3$X} \\
\midrule

25K
& 62.5 & 41.7 & 50.0
& \best{75.0} & 41.7 & \best{55.0}
& 37.5 & 25.0 & 30.0
& \best{50.0} & \best{50.0} & \best{50.0} \\

30K
& 75.0 & 58.3 & 65.0
& 75.0 & 58.3 & 65.0
& 37.5 & 41.7 & 40.0
& \best{62.5} & \best{58.3} & \best{60.0} \\
\midrule
\multicolumn{13}{l}{\textbf{Pouring Nut} \quad G$^3$VLA supervision: GT} \\
\midrule
25K
& 100.0 & 70.8 & 82.5
& 100.0 & \best{83.3} & \best{90.0}
& 93.75 & 66.7 & 77.5
& 93.75 & \best{70.8} & \best{80.0} \\

30K
& 100.0 & 75.0 & 85.0
& 100.0 & \best{87.5} & \best{92.5}
& 100.0 & 66.7 & 80.0
& 87.50 & \best{79.2} & \best{82.5} \\
\bottomrule
\end{tabular*}
\tablenote{Scores use the real-world evaluation protocol in
Sec.~\ref{sec:exp_setup}. ID: views 1 and 3; OOD: held-out views 11--13. Bold compares each G$^3$VLA variant with its corresponding backbone.}

\end{minipage}
\end{table*}

\textbf{Viewpoint generalization in real-world experiments.}
We next evaluate G$^3$VLA under unseen camera viewpoints in real world experiments. Models are trained using views 1--10 and evaluated on two training-distribution views (1 and 3) and three held-out views (11--13) (Table~\ref{tab:real_robot_res}). On Pouring Nut, G$^3$VLA consistently improves OOD performance for both $\pi_0$ and $\pi_{0.5}$. On Pick-and-Place Test Tube, it matches the $\pi_0$ baseline on OOD views while improving $\pi_{0.5}$ from 25.0\% to 50.0\% at 25K and from 41.7\% to 58.3\% at 30K. Overall, these results demonstrate that G$^3$VLA can improve generalization to physical camera shifts, with gains varying across tasks and backbones. \looseness=-1

\subsection{Calibration Attribution and Sensitivity}
\label{sec:calibration_controls}

We study whether the gains of G$^3$VLA arise from meaningful camera calibration and how sensitive the model is to calibration error. All controls use the same G$^3$VLA architecture and $\pi^3$X supervision, varying only the supplied calibration (Table~\ref{tab:calibration_controls}).

\textbf{Effect of camera calibration.}
With correct calibration, G$^3$VLA achieves 87.0\% and 97.0\% success
with $\pi_0$ and $\pi_{0.5}$, respectively.
Randomizing $K,T$ only at inference reduces success to 79.2\% and 90.6\%.
When the same calibration randomization is applied during both training
and evaluation, performance recovers to 83.8\% and 94.6\%, but remains
3.2 and 2.4 percentage points below the correctly calibrated models and
also below the corresponding vanilla VLA baselines (84.6\% and 95.9\%).
These controls indicate that the gains arise from meaningful camera
geometry rather than additional model capacity. For both randomization controls, one perturbation is sampled per rollout and applied consistently to the live camera poses throughout the rollout.

\textbf{Sensitivity to calibration error.}
G$^3$VLA is robust to moderate calibration noise. Perturbing focal length by $\pm2\%$, principal point by $\pm2$ pixels, camera translation by up to 1\,cm, and camera rotation by up to
$1^\circ$ reduces success by only 0.4 and 0.3 percentage points for $\pi_0$ and $\pi_{0.5}$, respectively. Together, these results show that G$^3$VLA benefits from geometrically
meaningful calibration without requiring exact calibration. \looseness=-1

\begin{table}[tbp]
\centering
\resulttablestyle
\setlength{\tabcolsep}{7pt}
\begin{lrbox}{\resulttablebox}
\begin{tabular}{@{}lcc@{}}
\toprule
\textbf{Calibration condition} & $\pi_0$ & $\pi_{0.5}$ \\
\midrule
Vanilla VLA baseline & 84.6 & 95.9 \\
\midrule
Correct $K,T$ & \textbf{87.0} & \textbf{97.0} \\
Small perturbation & 86.6 & 96.7 \\
Random $K,T$ (inference) & 79.2 & 90.6 \\
Random $K,T$ (training + inference) & 83.8 & 94.6 \\
\bottomrule
\end{tabular}
\end{lrbox}
\begin{minipage}{\wd\resulttablebox}
\caption{Calibration controls on LIBERO}
\label{tab:calibration_controls}
\usebox{\resulttablebox}
\tablenote{Macro-average success (\%). The four calibration conditions use the same G$^3$VLA modules and $\pi^3$X supervision within each backbone.}
\end{minipage}
\end{table}

\subsection{Ablation Studies}
\label{sec:ablation_studies}

\begin{figure}[t]
    \centering
     \includegraphics[width=0.98\linewidth]{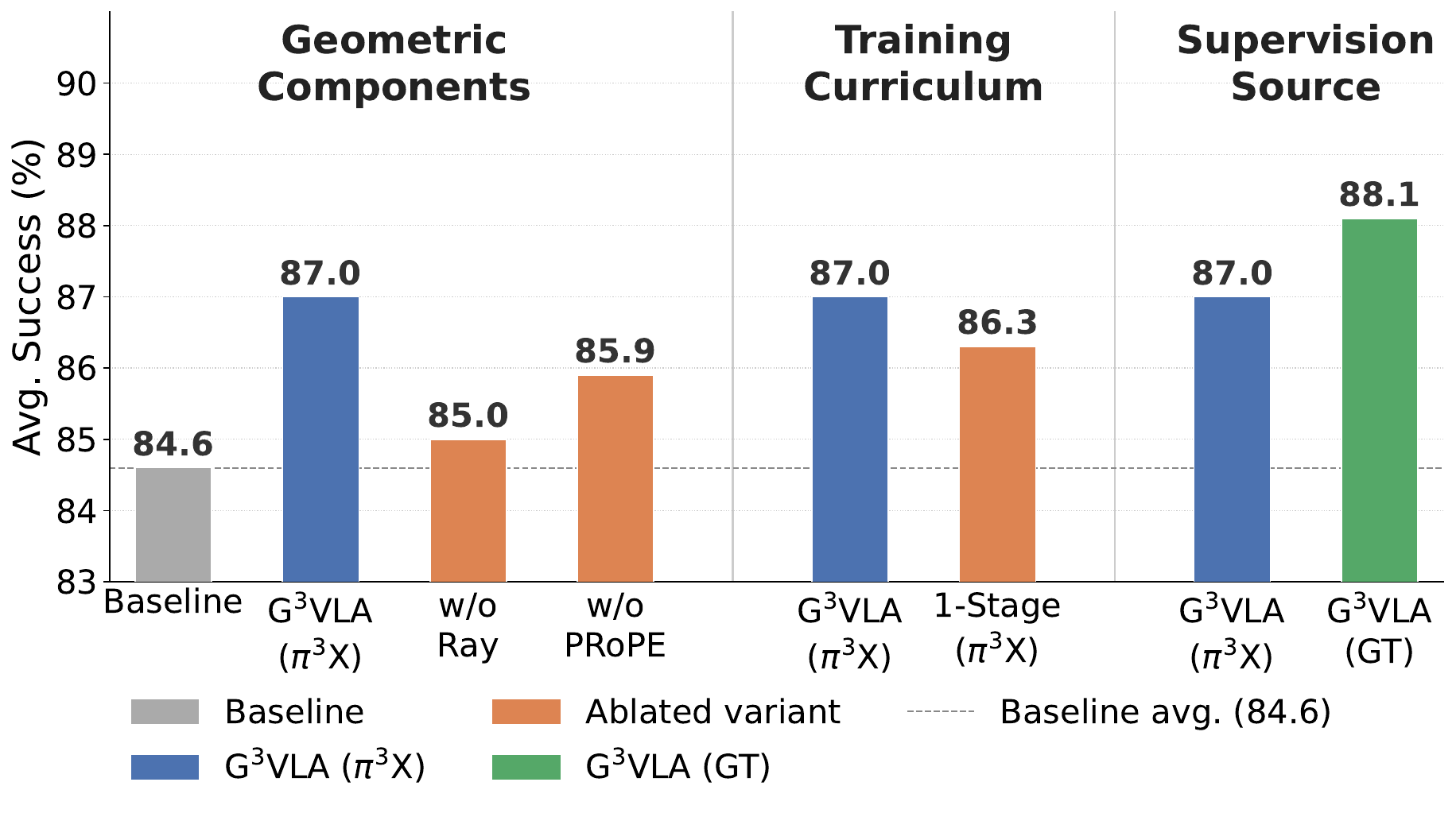}
    \caption{\textbf{Ablation study on LIBERO ($\pi_0$).}
    We study geometric components, training curriculum, and supervision source. The gray dashed line denotes the vanilla VLA baseline (84.6\%).}
    \label{fig:ablate}
\end{figure}

We ablate the geometric components, training curriculum, and supervision source of G$^3$VLA on LIBERO with $\pi_0$ (Fig.~\ref{fig:ablate}).
Unless otherwise specified, all variants use $\pi^3$X supervision.
Removing ray embeddings while preserving cross-view fusion reduces success from 87.0\% to 85.0\%, while removing PRoPE but retaining ray embeddings reduces it to 85.9\%. 
Replacing the two-stage curriculum with one-stage joint training reduces the success rate from 87.0\% to 86.3\%.
Finally, using GT geometry increases success from 87.0\% with $\pi^3$X supervision to 88.1\%, showing that $\pi^3$X retains most of the benefit of GT geometry without requiring GT geometry during training.

We additionally study sensitivity to the confidence threshold $\tau$ in
Table~\ref{tab:confidence_threshold}. Success remains stable at
86.8--87.0\% for $\tau=0.05$--$0.20$, despite the fraction of retained
pixels decreasing from 94.11\% to 73.66\%. More aggressive filtering
reduces both supervision coverage and success. 

\begin{table}[!htbp]
\centering
\resulttablestyle
\setlength{\tabcolsep}{3pt}

\begin{lrbox}{\resulttablebox}
\begin{tabular}{@{}lcccccc@{}}
\toprule
$\tau$ & 0.05 & \textbf{0.10} & 0.20 & 0.30 & 0.40 & 0.50 \\
\midrule
Retained (\%) & 94.11 & 88.31 & 73.66 & 48.26 & 24.95 & 8.75 \\
Success (\%) & 86.8 & \textbf{87.0} & 86.9 & 86.4 & 85.8 & 85.1 \\
\bottomrule
\end{tabular}
\end{lrbox}
\begin{minipage}{\wd\resulttablebox}
\caption{Confidence-threshold sensitivity}
\label{tab:confidence_threshold}
\usebox{\resulttablebox}
\tablenote{Retained: percentage of pixels passing the confidence gate.}

\end{minipage}
\end{table}

\subsection{Generalization Across Backbones}

G$^3$VLA also generalizes beyond the $\pi_0$ backbone
(Table~\ref{tab:backbone_generalization}). On $\pi_{0.5}$, G$^3$VLA with $\pi^3$X
supervision improves the reproduced baseline from 95.9\% to 97.0\%.
On GR00T~1.5, whose two-tower architecture exposes visual features to
the diffusion policy through cross-attention to a frozen VLM,
G$^3$VLA~($\pi^3$X) improves the macro average from 94.9\% to 95.3\%.  \looseness=-1

\begin{table}[tbp]
\centering
\resulttablestyle
\setlength{\tabcolsep}{3pt}
\begin{lrbox}{\resulttablebox}
\begin{tabular}{@{}llccccc@{}}
\toprule
\textbf{Backbone} & \textbf{Variant}
& Spatial & Object & Goal & L-10 & \textbf{Avg.} \\
\midrule
$\pi_{0.5}$
& Baseline (reprod.)
& \best{98.8} & 98.4 & 94.4 & 91.8 & 95.9 \\

& G$^3$VLA ($\pi^3$X)
& 98.2 & \best{99.0} & \best{98.0} & \best{92.8} & \best{97.0} \\
\midrule
GR00T~1.5
& Baseline
& \best{96.6} & 97.0 & 95.4 & 90.6 & 94.9 \\

& G$^3$VLA ($\pi^3$X)
& \best{96.6} & \best{99.0} & \best{95.8} & 89.6 & \best{95.3} \\
\bottomrule
\end{tabular}
\end{lrbox}

\begin{minipage}{\wd\resulttablebox}
\caption{LIBERO performance across VLA backbones.}
\label{tab:backbone_generalization}
\usebox{\resulttablebox}
\tablenote{Success rates (\%). Bold compares variants within each backbone.}
\end{minipage}
\end{table}

\subsection{Computational Cost}
G$^3$VLA adds 56.7M parameters
(1.62\%) to $\pi_0$ during training and 48.3M (1.38\%) at deployment
after removing the auxiliary head. On RoboCasa using the same L40S GPU,
inference latency just slightly increases from 112.3 to 115.7\,ms (+3.0\%).
Generating the LIBERO $\pi^3$X cache requires 1.94 GPU-hours on one L40S,
while Stage~1 training takes 1.36 hours on four H200 GPUs
(5.46 GPU-hours) with 19.4\,GB peak memory per GPU. \looseness=-1

\section{Conclusion}
\label{sec:conclusion}
We presented G$^3$VLA, a visual-token interface that injects explicit
camera calibration into pretrained VLA policies without changing their
action representation. Across LIBERO, RoboCasa24, RoboTwin2.0, and
real-robot evaluations, G$^3$VLA improves manipulation performance and
shows particularly strong gains under held-out camera viewpoints.
Matched calibration controls further indicate that these gains arise
from meaningful camera geometry rather than additional model capacity
alone. Overall, our results show that explicit camera calibration provides
an effective geometric prior for improving viewpoint robustness in
pretrained VLAs.  \looseness=-1

\FloatBarrier

\FloatBarrier
\bibliographystyle{IEEEtran}
\bibliography{references}  %

\end{document}